\documentclass{article}
\usepackage{spconf,amsmath,graphicx}

\usepackage{bm}
\usepackage{multirow}
\usepackage{amsmath}
\usepackage{amsfonts}
\usepackage{algorithm} 
\usepackage{tipa}
\usepackage{algpseudocode} 
\usepackage{url}

\usepackage{breakurl}
\usepackage[breaklinks]{hyperref}
\usepackage{cite}
\usepackage{xcolor}

\def\R{{\mathbb R}}
\def\nlang{{70}}
\PassOptionsToPackage{hyphens}{url}\usepackage{hyperref}
\hypersetup{colorlinks=true}

\title{Massively Multilingual ASR on 70 Languages: \\Tokenization, Architecture, and Generalization Capabilities}
\name{ \begin{tabular}{c}Andros Tjandra, Nayan Singhal, David Zhang, \\ Ozlem Kalinli, Abdelrahman Mohamed, Duc Le, Michael L. Seltzer \end{tabular}}
\address{ Meta AI, USA  \\
{\small \tt  androstj@meta.com} 
}
\begin{document}
\ninept
\maketitle
\begin{abstract}

End-to-end multilingual ASR has become more appealing because of several reasons such as simplifying the training and deployment process and positive performance transfer from high-resource to low-resource languages. However, scaling up the number of languages, total hours, and number of unique tokens is not a trivial task. This paper explores large-scale multilingual ASR models on \nlang{} languages. We inspect two architectures: (1) Shared embedding and output and (2) Multiple embedding and output model. In the shared model experiments, we show the importance of tokenization strategy across different languages. Later, we use our optimal tokenization strategy to train multiple embedding and output model to further improve our result. Our multilingual ASR achieves 13.9\%-15.6\% average WER relative improvement compared to monolingual models. We show that our multilingual ASR generalizes well on an unseen dataset and domain, achieving 9.5\% and 7.5\% WER on Multilingual Librispeech (MLS) with zero-shot and finetuning, respectively.
\end{abstract}
\begin{keywords}
Multilingual, speech recognition, multi-softmax, RNN-T, tokenization
\end{keywords}
\vspace{-0.3cm}

\section{Introduction}
\label{sec:intro}

Modern end-to-end (E2E) automatic speech recognition (ASR) models, such as recurrent neural network transducers (i.e., RNN-T)~\cite{Graves12,GravesMohamedHinton13,RaoSakPrabhavalkar17,HeSainathPrabhavalkarEtAl19,YehMahadeokarKalgaonkar19,ZhangLuSakEtAl20,ShiWangWuEtAl21} and attention-based encoder-decoder models~\cite{ChanJaitlyLeEtAl16,BahdanauChorowskiSerdyukEtAl16,KimHoriWatanabe17,ChiuSainathWeEtAl18,ZeyerIrieSchluterEtAl18}, 
have the ability to benefit from very large amounts of speech data~\cite{zheng22d_interspeech}. They capture acoustic variabilities in their encoder modules while modeling the linguistic structure of language in their decoder modules.
Due mainly to this scalability, E2E models have enabled great strides in multilingual ASR research, which aims to create a single model that can recognize multiple languages at the same time~\cite{Li18multidialect,kannan19_interspeech,Hou2020LargeScaleEM, Pratap2020,li2021scaling,babu22_interspeech,yang2022learning,joshi21_interspeech}.

Training and deploying multilingual ASR models has several practical benefits. A single well-performing multilingual model saves significant manual effort to tune each language individually during training, as well as model maintenance and deployment when in production. Building a highly accurate large-scale multilingual teacher model also enables scaling up semi-supervised training to more languages via iterative pseudo-labeling. A single multilingual model ensures seamless deployment and good ASR quality for locales with heterogeneous languages as it integrates language identification signals and speech recognition signals during decoding. In some cases, there is a positive effect where high-resource languages also improve low resource languages performance when we mix and train them simultaneously.

One key challenge of multilingual ASR in practice is when we scale up the number of languages, our vocabulary size grows larger. Prior works \cite{li2021scaling} trained a multilingual ASR for 15 languages by simply combining all graphemes together. \cite{joshi21_interspeech} proposed multilingual with multi-decoder output for 4 languages. \cite{Pratap2020} built multilingual ASR on top of 50 languages with total 16,000 hours training dataset. To the best of our knowledge, most of the existing works have less amount of languages and smaller unique graphemes size compared to our in-house video dataset.

In this paper, we explore large-scale multilingual on \nlang{} languages with 150,000 hours dataset. We explored based on two architectures: shared input embedding and output architecture and multiple input embedding and output architecture. In the shared model experiment, we show the importance of tokenization strategy across different languages and provide further analysis of the result. Later, we use the best token strategy to train a multilingual model with multiple input embedding and output layers. We evaluate our multilingual model on in-house dataset and show significant improvement over monolingual models. Lastly, we also show our multilingual model could generalized well on the new domain and dataset. We achieve 9.5\% and 7.5\% WER on Multilingual Librispeech (MLS) with zero-shot and finetuning, respectively. To the best of our knowledge, our result is competitive with the state-of-the-art performance on MLS dataset.

\section{Multilingual ASR}
\begin{figure*}[t]
    \centering
    \includegraphics[width=0.68\textwidth]{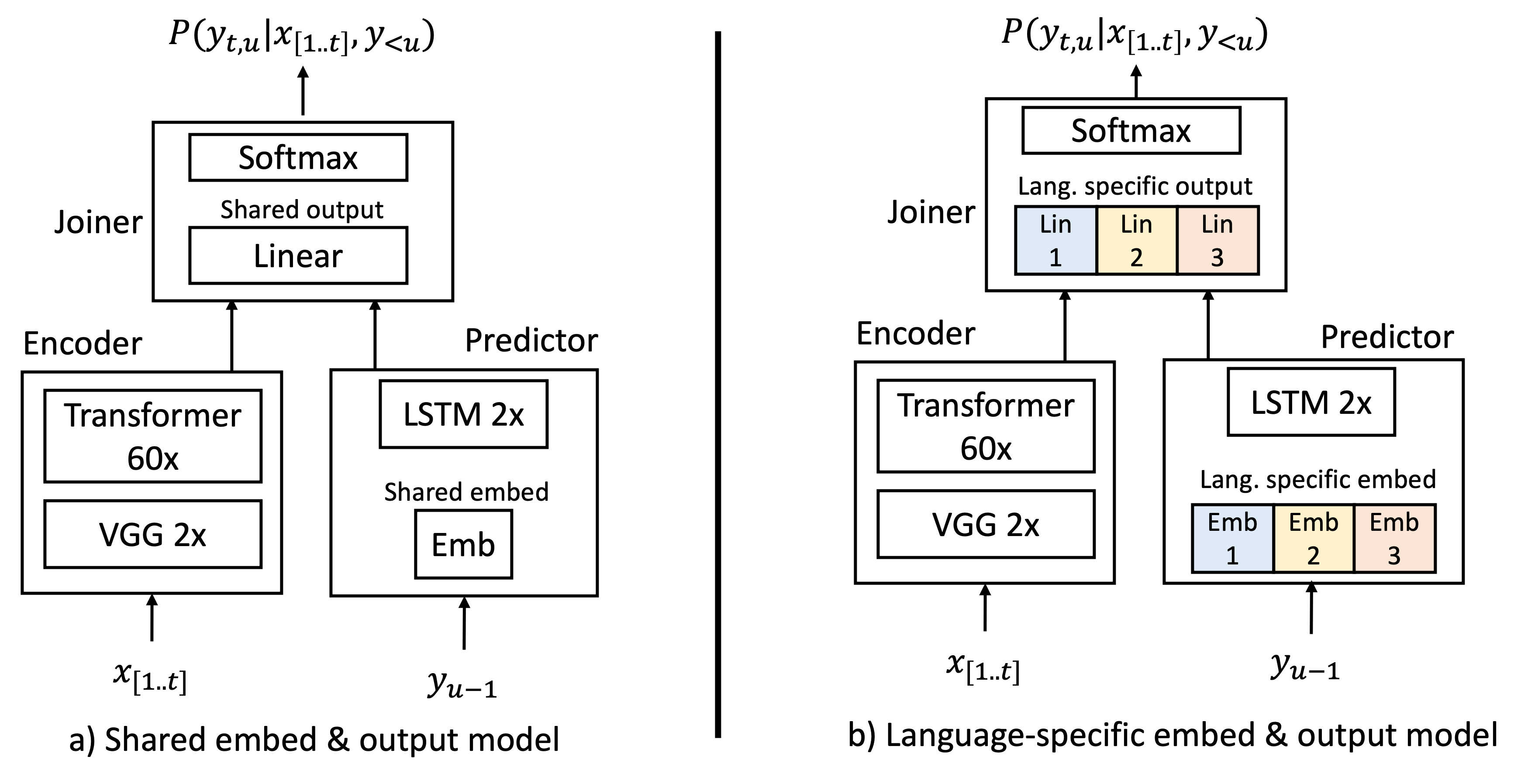}
    \caption{Multilingual model with a) shared input embedding and output linear layer; b) multiple input embedding and output linear layer. }
    \label{fig:shared_lang_rnnt}
    \vspace{-0.3cm}
\end{figure*}
\subsection{Model Architecture}
Our multilingual model is based on an end-to-end Transducer model \cite{graves2012sequence} that is composed by encoder, predictor, and joiner modules. Let $\mathbf{x} = [x_1,..,x_T]$ be an input speech features with length $T$ and $\mathbf{y}=[y_1,...y_U]$ be an output token sequence with length $U$. The encoder model process $\mathbf{x}$ and produce higher level acoustic representation $\mathbf{h^{enc}}=[h_1^{enc},..,h_T^{enc}]=enc(\mathbf{x})$. The prediction network is (usually) an autoregressive decoder that produces hidden states $\mathbf{h^{pred}} = [h_1^{pred},..,h_U^{pred}]$. Each hidden states $h_{u}^{pred} = pred(y_{1},..,y_{u-1})$ conditioned on previous output tokens $y_{<u}$. The joiner module combines both encoder $h_t^{enc}$ and predictor $h_u^{pred}$ representation to calculate the logits $z_{t,u} = joiner(h_t^{enc}, h_u^{pred})$. Lastly, we apply softmax function to calculate the probability given input and previous output tokens $P(y_{t,u} | \mathbf{x}_{[1..t]}, y_{<u}) = softmax(z_{t,u})$. Transducer loss is defined as a negative log-likelihood of output sequence given the input features $\mathcal{L(\mathbf{x}, \mathbf{y}; \theta)}=-\log P_{\theta}(\mathbf{y}|\mathbf{x})$ where $\theta = \{enc, pred, joiner\}$ parameters.

Here, our encoder module starts with three VGG layers (where each VGG layer consisted of two 2d-convolution layers with kernel size = 3 and pooling size = 2) to reduce the speech feature length by 8 times and followed by 60 Transformer \cite{Vaswani2017AttentionIA} layer with input dimension 1152, feedforward dimension = 4608, head = 16 and GeLU activation function \cite{Hendrycks2016GaussianEL}. We set dropout $p=0.1$ for every Transformer layers. The predictor consisted of an input embedding layer with output dimension = 1024, 2 LSTM \cite{Hochreiter1997LongSM} layers with hidden dimension = 512 with dropout with $p=0.3$ and layer normalization \cite{Ba2016LayerN}. The joiner layers combined the encoder and predictor hidden representation by elementwise sum operation, followed by tanh activation and a softmax linear layer. In this work, we investigate two types of Transducer models: (1) shared input embedding and output architecture, (2) language-specific multiple input embedding, and output linear architecture. 

\subsection{Shared Input Embedding and Output Architecture}
One of the simplest ways to tackle multilingual ASR is by having a shared representation between all languages. In Figure \ref{fig:shared_lang_rnnt}a, we construct a Transducer with single input embedding and output layer to represent all languages. As we know Transducer loss requires high memory usage, specifically $O(B \times T \times U \times V)$ complexity, which scales linearly with vocabulary sizes. This motivates us to find a different way to represent each language's transcription to achieve a better tradeoff between efficiency and model accuracy.

Several representation choices exist depending on the language and their tokenization or pre-processing. Some languages like English, French, etc... (Indo-European languages) with Latin-based alphabet have small unique characters and could be represented as characters or sub-words \cite{kudo2018subword}. Some research has shown that using sub-words provides better results compared to phonemes or characters \cite{irie19_interspeech}. On the other hand, some languages (e.g., Mandarin, Japanese, Korean) have very large unique characters and it is not feasible to use sub-words to represent them. To reduce the size, one could represent those languages with their phonemes (e.g., Mandarin Pinyin, Korean Jamo). However, \cite{zhou2018comparison, Wang2020} showed that using phonemes hurts the performance and requires extra post-processing steps (e.g., WFST \cite{Mohri2002WeightedFT}) to convert the model output back to their original written form transcriptions.

Let $\cup$ be the set-union operator that merges two sets and keeps only unique elements inside. We explore two approaches: 
 \begin{enumerate}
    \item Let $\mathcal{L} = [1...L]$ be a set of $L$ languages. For all languages $l \in \mathcal{L}$, we extract characters $V_{l}$ = extract\_char(language $l$) and union all into $\mathcal{V}_{char} = \left( \bigcup_{l \in \mathcal{L}} \mathcal{V}_{l} \right)$.
    \item For all languages $o \in \mathcal{O} \subseteq \mathcal{L}$ with unique characters larger than 512, we set $\mathcal{V}_{o}$ = extract\_char(language $o$). For all languages $p \in \mathcal{P} \subseteq \mathcal{L}$ with unique character less than 512, we extract subword with max. 512 tokens $\mathcal{V}_{p}$ = extract\_subw(language $p$, size=512). Lastly, we union all subwords and characters token into $\mathcal{V}_{char+subw} = \left( \bigcup_{o \in \mathcal{O}} \mathcal{V}_{o} \right) \cup \left( \bigcup_{p \in \mathcal{P}} \mathcal{V}_{p} \right)$.
\end{enumerate}
The first approach is the simplest one where we just use basic tokenization to get all characters from each language and union all of them by removing any duplicates to achieve minimum vocabulary size. The second approach takes advantage of subword tokenization that has been proven to work well in end-to-end ASR \cite{irie19_interspeech}. By combining the subword tokenization for languages with small unique characters and characters tokenization for some languages with a large number of unique characters, we aim to strike a balance between minimizing the vocabulary size and improving the ASR performance.
\begin{figure*}[t]
    \centering
    \includegraphics[width=1.0\textwidth]{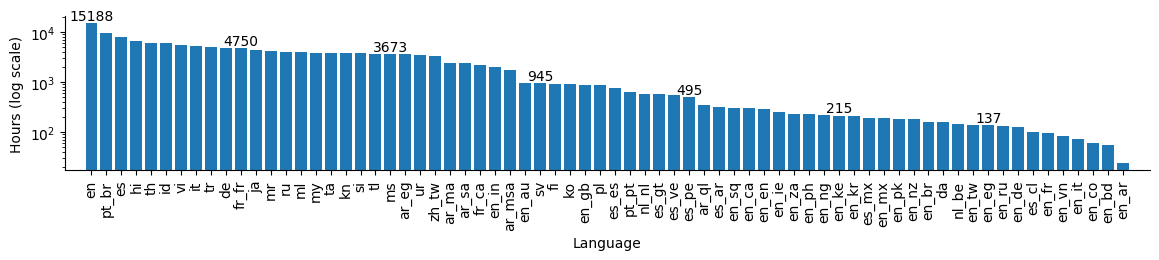}
    \caption{Language vs. hours of training data for \nlang{} languages.}
    \label{fig:plot_hours}
\end{figure*}

\subsection{Multiple Embedding and Output Architecture}
We explore another architecture with language-specific input embedding and output linear layer. Figure \ref{fig:shared_lang_rnnt}b shows the architecture differences compared to the shared language architecture. Let $d_{emb}$ be the embedding output dimension, $d_{hid}$ be the LSTM hidden dimension, and $\hat{\mathcal{V}}_{l}=\mathcal{V}_{l} \cup \O$ be the tokens for language $l$ plus a blank token $\O$. Then we have multiple embedding weights $\mathbf{E} = [E_1,..E_L]$ where $ E_{l} \in \R^{|\hat{\mathcal{V}}_{l}| \times d_{emb}}$ and multiple output linear weights $\mathbf{W} = [W_1,..,W_{L}]$ where $ W_{l} \in \R^{d_{hid} \times |\hat{\mathcal{V}}_{l}|}$ and $\mathbf{b} = [b_1,..,b_{L}]$ where $ b_{l} \in \R^{|\hat{\mathcal{V}}_{l}|}$. For each language input embedding and output layer, we use characters if that language contains $>512$ unique characters, otherwise, we use subwords with 512 tokens. One advantage of this language-specific architecture is that we could represent the same token between different languages with different embedding and weight matrices. Thus, we could disambiguate characters and subwords that look the same in the written space but sound different (e.g., `a' is spelled as `\textipa{eI}' in English and `\textipa{a}' in Indonesian).

\section{Experimental Setup}

\subsection{Dataset}
We conduct the experiments on our in-house datasets. All in-house datasets are de-identified public videos with no personally identifiable information (PII). Figure \ref{fig:plot_hours} shows the amount of hours training data for each language. Overall, we have around 150,000 hours across \nlang{} languages.

We run data alignment and segmentation pipelines to remove long silences or low-quality transcription and segment original speech and transcripts into maximum of 10-second chunks.  To increase the amount of training data, we apply speed perturbation \cite{ko15_interspeech} with speed factors of 0.9x and 1.1x. 
Since there are some languages with low amount of data, we apply batch re-sampling \cite{babu22_interspeech} by sampling from $p_l \sim \left( \frac{n_l}{N}\right)^{\alpha}$ with $\alpha = 0.5$.
We do not apply any extra processing for the transcript and keep all casings and punctuations for our model to learn them in an end-to-end fashion.

For the shared model experiment, we prepare several vocabulary sets. The character-only vocabulary set $\mathcal{V}_{char}$ consists of 11320 unique tokens after unionization. The majority of the tokens come from Mandarin (8542 unique characters), Japanese (4177 unique characters), and Korean (2313 unique characters). The character+subword vocabulary set $\mathcal{V}_{char+subw}$ consists of 18744 unique tokens, which is the union of Mandarin, Japanese, Korean characters, and 512 subwords for each of the remaining languages. For the language-specific model experiment, we use characters token to represent Mandarin, Japanese, and Korean, and 512 subwords for each of the remaining languages.

We evaluate our model with test data on two different domains, \textit{vid-clean} and \textit{vid-noisy}, for every language. The main difference between these two domains is that the data from \textit{vid-noisy} are more acoustically challenging compared to \textit{vid-clean}. The amount of data for each language ranges from 5-50 hours for \textit{vid-clean} and 12-50 hours for \textit{vid-noisy}.

\subsection{Training Details}
We extract 80-dimension log-Mel spectrogram features with step-size 10ms and window-size 25ms from the audio waveform. We apply mean-variance normalization for each time-step and SpecAugment \cite{park19e_interspeech} with $W=80$, $F=27$, $m_F=1$, $T=100$, $p=1.0$, $m_T=2$.

Our multilingual models have around 1 billion parameters for all scenarios (with little variation depending on the vocabulary size). To improve training throughput and memory efficiency, we use several tricks following \cite{zheng22d_interspeech} such as: fully sharded data parallel (FSDP) \cite{FairScale2021}, activation checkpointing \cite{chen2016training} and mixed-precision training \cite{micikevicius2017mixed}. We use Adam optimizer \cite{kingma2014adam} with peak $lr=4e-4, \beta_1=0.9, \beta_2=0.98$. We train our model for up to 700,000 updates with 64 GPUs. We apply $lr$ warm-up for the first 20,000 updates and exponentially decay for the rest of the remaining updates into $0.1$ from the peak $lr$.

\section{Results}

\begin{figure*}[t]
    \centering
    \includegraphics[width=1\textwidth]{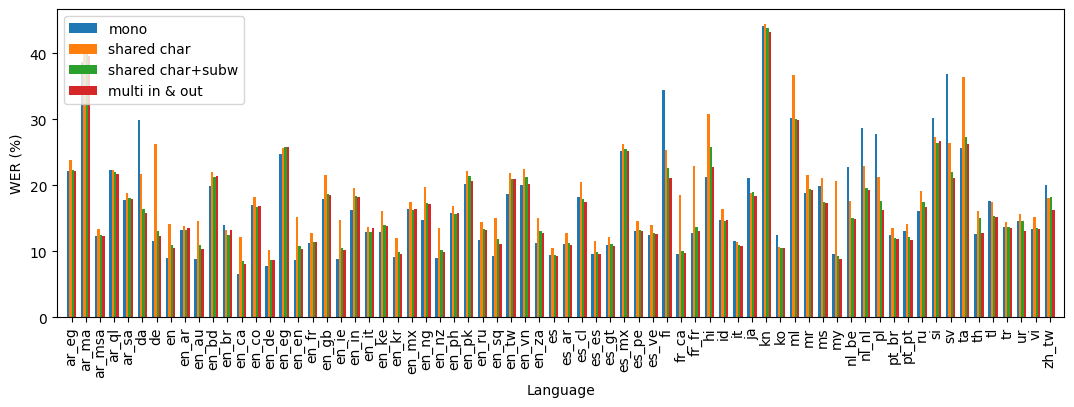}
    \caption{WER comparison across all languages on in-house dataset test \textit{vid-clean}.}
    
    \label{fig:plot_cer_noisy}
\end{figure*}
In this section, we discuss our experimental results in different scenarios and followed by a discussion. Our monolingual baselines are 100 million parameters Emformer-based RNN-T  \cite{ShiWangWuEtAl21}. For some monolingual baselines, we combine groups of languages with high similarity together (e.g., pt-pt and pt-br) for improving the performance. We use word error rate (WER) for most of the word-based languages and character error rate (CER) for character-based languages (i.e. Japanese (ja), Mandarin (zh-tw), Korean (ko), Burmese (my) and Thai (th)).

\subsection{Results on In-House Dataset}
\begin{table}[ht]
\vspace{-0.3cm}
\centering
\caption{Average WER(\%) on test \textit{vid-clean} and \textit{vid-noisy}.}
\begin{tabular}[t]{ l  c  c }
 \hline
 \textbf{Model} & \textbf{\texttt{vid-}} & \textbf{\texttt{vid-}} \\ 
 & \textbf{\texttt{clean}} & \textbf{\texttt{noisy}} \\ \hline \hline
 Monolingual {(MO)}& 19.2 & 20.2 \\ 
 \hline
 Multilingual &  &  \\  
  \quad w/ shared char {(ML-SC)} & 20.6 & 21.4 \\
  \quad w/ shared char + subw. {(ML-SCW)}& 16.9 & 17.8 \\
  \quad w/ lang. spec. in\&out  {(ML-IO)}& 16.2 & 17.4 \\
 \hline
\end{tabular}
\vspace{-0.3cm}
\label{tab:summary}
\end{table}

We report our results on in-house datasets \textit{vid-clean} and \textit{vid-noisy} in Table \ref{tab:summary}. The multilingual model with shared character strategy {ML-SC} shows inferior results with relative WER degradation of 5.9\%-7.3\% compared to {MO}. By using the shared char+subword strategy, {ML-SCW} shows relative WER improvement of 16.8\%-18\% compared to {ML-SC} and 11.9\%-12\% compared to {MO}. On top of that, by adding language-specific embedding and output layer, {ML-IO} achieved relative WER improvement of 2.2\%-4.1\% compared to {ML-SCW} and 13.9\%-15.6\% compared to {MO}. 

Figure \ref{fig:plot_cer_noisy} shows the detailed comparison for all languages between all models. As we can see, our multilingual models {ML-SWC} and {ML-IO} gain significant improvement especially on low resource languages (e.g., Danish (da), Polish (pl)).

\subsection{Analysis on Shared Character vs. Shared Character + Subword}
We observe a relatively sizeable performance gap between shared char and shared char + subword model in the previous section. Our hypothesis is different tokenization strategies significantly change the number of decoding steps for each language. We calculate the statistic of the number of tokens per second after we apply tokenization to the transcription. Then, we got $11.5 \pm 2.6, \max=15.3, \min=3.8$ on shared char, and $4.8\pm0.7, \max=6.8, \min=3.5$ on shared char+subword. As we can see, shared char tokenization leads to longer decoding time-step and higher variance between different languages. 

To confirm our hypothesis, we dive deeper by comparing the inference result. First, we choose Tamil ({ta}) result as we observed the largest performance gap between ML-SC and ML-SCW. ML-SC gets 36.5\% WER with 3.5\% insertion, 11.5\% deletion, 21.4\% substitution error rate and ML-SCW gets 27.3\% WER with 5\% insertion, 3.5\% deletion, 18.7\% substitution error rate. We could a big difference in the deletion error rate, but their insertion and substitution error rates remain similar. Coincidently, Tamil has the highest number of tokens per second (15.3). On the other hand, ML-SC could match or improve ML-SCW performance in some languages (i.e. Japanese, Korean) with similar tokens per second.

Lastly, we calculate the error rate on all languages and we found ML-SC is averaging 4.8\% insertion, 6.7\% deletion, 9.9\% substitution, and ML-SCW is averaging 5\% insertion, 3.7\% deletion, 9.1\% substitution error rate. We observed the same thing happening where the deletion error rate increased by 44\% by using shared char instead of shared char+subword tokenization. We conclude that simply using shared character tokenization could lead the multilingual model to subpar performance due to the high variation of decoding time-step between different languages. We show that by minimizing the variance of decoding steps between languages with clever combinations between subwords and characters, we could significantly improve our multilingual result.

\subsection{Results on MLS: Zero-shot and Finetuning}
Here, we want to observe how our multilingual model generalizes on new unseen domains and datasets. Therefore, we perform a zero-shot task to see how our model performs directly without any additional training steps and adaptation. After that, we also want to measure how good our model is to be adapted into a new dataset by using it as a seed model. For this purpose, we use Multilingual Librispeech (MLS) \cite{pratap2020mls} dataset that consisted of 8 languages (English, German, Dutch, French, Spanish, Italian, Portuguese, and Polish). Table \ref{tab:mls} shows MLS average WER from prior works and our experiment. We achieve 9.5\% WER on zero-shot and 7.5\% WER after finetuning. To the best of our knowledge, our result is competitive with the state-of-the-art performance on MLS dataset.
\begin{table}[ht]
\centering
\vspace{-0.3cm}

\caption{Average WER(\%) on MLS test sets.}
\begin{tabular}[t]{ l  c }
 
 \hline
 \textbf{Model} & \textbf{WER (\%)} \\ \hline \hline
 
 \textbf{Prior works}  \\ 
 Monolingual CTC \cite{Pratap2020} & 11.8 \\
 Monolingual CTC w/ 5-gram LM \cite{Pratap2020} & 10.7 \\
 XLS-R (ft. 10h) \cite{babu22_interspeech} & 13.8 \\
 RNN-T 1B (ft. all) \cite{li2021scaling} & 7.9 \\
 \hline
 \textbf{Zero-shot} \\
 {MO} & 13.7 \\ 
 {ML-SCW} & 9.8  \\  
 {ML-IO} & 9.5  \\ 
 \hline
 \textbf{Finetune all}  \\  
 {ML-SCW} & 7.7  \\  
 {ML-IO} & 7.5  \\ 
 \hline
\end{tabular}
\label{tab:mls}
\end{table}

\section{Conclusion}
In this paper, we conducted a study about large-scale multilingual ASR across \nlang{} languages and on around 150,000 hours of speech data. We show that balancing the tokenization strategy is very important to achieve better WER on both word-based and character-based languages. Our proposed multilingual ASR with char+subword achieves 12\% WER improvement compared to monolingual models. On top of that, by adding language-specific embedding and output layer, we achieve 13.9-15.6\% WER improvement over the monolingual model. We also show our model could generalize and adapt to unseen domains and datasets. We achieve 9.5\% and 7.5\% WER on Multilingual Librispeech (MLS) with zero-shot and finetuning, respectively. In the future, we plan to scale up the amount of training data by adding pseudo-labeling pipeline in every language.
\vfill\pagebreak

\footnotesize
\bibliographystyle{IEEEbib}
\bibliography{refs}

\end{document}